\documentclass[10pt,twocolumn,letterpaper]{article}

\PassOptionsToPackage{table}{xcolor}

\usepackage[pagenumbers]{cvpr} 

\usepackage{cuted} 

\usepackage{currfile} 

\usepackage{caption} 

\usepackage{algorithm}
\usepackage{algorithmic}
\usepackage{multirow}
\usepackage{array}
\usepackage{longtable} 
\usepackage[normalem]{ulem} 
\usepackage{tikz}
\usetikzlibrary{positioning,fit,calc,arrows.meta,backgrounds}
\tikzset{
  arch/.style={font=\scriptsize, >={Stealth[length=4pt]}},
  mllm/.style={draw=violet!65!black, fill=violet!9, rounded corners=2.5pt, line width=0.6pt, align=center, inner sep=3pt},
  code/.style={draw=black!50, fill=black!4, rounded corners=2.5pt, line width=0.6pt, align=center, inner sep=3pt},
  datum/.style={draw=blue!45!black, fill=blue!4, rounded corners=1pt, line width=0.5pt, align=center, inner sep=2.5pt},
  outbox/.style={draw=green!40!black, fill=green!8, rounded corners=2.5pt, line width=0.6pt, align=center, inner sep=3pt},
  optional/.style={dashed, dash pattern=on 2.2pt off 1.4pt},
  flow/.style={->, line width=0.7pt, black!60},
  mflow/.style={->, line width=0.75pt, violet!65!black},
  oflow/.style={->, line width=0.7pt, black!55, dashed, dash pattern=on 2.2pt off 1.4pt},
  panel/.style={draw=black!18, fill=black!1.8, rounded corners=4pt, line width=0.5pt},
  ptitle/.style={font=\scriptsize\bfseries, text=black!72, anchor=north west, inner sep=2.5pt},
  anote/.style={font=\tiny, text=black!62, align=left, inner sep=1pt},
  warn/.style={font=\tiny, text=red!60!black, align=left, inner sep=1pt},
  badge/.style={circle, fill=black!72, text=white, font=\tiny\bfseries, inner sep=0.4pt, minimum size=7.5pt},
  rcallout/.style={circle, fill=red!78!black, text=white, font=\scriptsize\bfseries, inner sep=0pt,
    minimum size=11pt, draw=white, line width=0.7pt},
}

\definecolor{icInk}{HTML}{4A4945}
\definecolor{icCond}{HTML}{2F6DB5}
\definecolor{icPlug}{HTML}{D9822B}
\definecolor{icEval}{HTML}{3C8D4F}
\definecolor{icMllm}{HTML}{7A4FB3}
\definecolor{icRed}{HTML}{D62728}
\tikzset{
  ic-mllm/.pic={
    \draw[icMllm, fill=icMllm!12, line width=0.6pt, rounded corners=1.5pt]
      (-0.28,-0.12) -- (-0.28,0.22) -- (0.28,0.22) -- (0.28,-0.12) -- (0.0,-0.12) -- (-0.14,-0.27) -- (-0.11,-0.12) -- cycle;
    \foreach \x in {-0.13,0,0.13} \fill[icMllm] (\x,0.05) circle (0.035);},
  ic-gear/.pic={
    \foreach \a in {0,45,...,315} \fill[icInk, rotate=\a] (-0.05,0.17) rectangle (0.05,0.25);
    \fill[icInk] (0,0) circle (0.19); \fill[white] (0,0) circle (0.08);},
  ic-compiler/.pic={
    \draw[icCond, fill=white, line width=0.55pt] (-0.22,-0.27) -- (-0.22,0.27) -- (0.12,0.27) -- (0.22,0.17) -- (0.22,-0.27) -- cycle;
    \foreach \y/\l in {0.14/0.3,0.04/0.3,-0.06/0.22,-0.16/0.16} \draw[icCond!80, line width=0.7pt] (-0.15,\y) -- ++(\l,0);
    \foreach \a in {0,60,...,300} \fill[icInk, rotate around={\a:(0.17,-0.19)}] (0.15,-0.1) rectangle (0.19,-0.06);
    \fill[icInk] (0.17,-0.19) circle (0.1); \fill[white] (0.17,-0.19) circle (0.04);},
  ic-vdm/.pic={
    \foreach \d/\c in {0.12/12,0.06/22} \draw[icInk, fill=black!\c, line width=0.4pt] (-0.28+\d,-0.18+\d) rectangle (0.2+\d,0.12+\d);
    \draw[icInk, fill=white, line width=0.55pt] (-0.28,-0.18) rectangle (0.2,0.12);
    \fill[icPlug!70] (-0.24,-0.15) -- (-0.1,0.02) -- (0.0,-0.06) -- (0.08,0.04) -- (0.17,-0.15) -- cycle;
    \fill[icPlug] (0.08,0.07) circle (0.03);},
  ic-tedio/.pic={
    \draw[black!45, line width=0.5pt] (-0.28,-0.05) -- (-0.2,0.12) -- (-0.12,-0.14) -- (-0.04,0.1) -- (0.04,-0.12) -- (0.12,0.14) -- (0.2,-0.08) -- (0.28,0.06);
    \draw[icPlug, line width=1.0pt] (-0.28,-0.02) .. controls (-0.1,0.06) and (0.1,-0.06) .. (0.28,0.03);},
  ic-bachvid/.pic={
    \foreach \x/\f in {-0.2/icPlug!55,0.0/white,0.2/white} \draw[icInk, fill=\f, line width=0.45pt] (\x-0.08,-0.2) rectangle (\x+0.08,-0.02);
    \draw[icPlug, line width=0.6pt, -{Stealth[length=2.3pt]}] (-0.2,-0.0) .. controls (-0.15,0.18) and (-0.05,0.18) .. (0.0,0.0);
    \draw[icPlug, line width=0.6pt, -{Stealth[length=2.3pt]}] (-0.2,-0.0) .. controls (-0.12,0.3) and (0.12,0.3) .. (0.2,0.0);},
  ic-memory/.pic={
    \draw[icInk, fill=icPlug!15, line width=0.5pt] (-0.2,-0.2) -- (-0.2,0.15) arc (180:360:0.2 and 0.07) -- (0.2,-0.2) arc (360:180:0.2 and 0.07);
    \draw[icInk, fill=icPlug!35, line width=0.5pt] (0,0.15) ellipse (0.2 and 0.07);
    \draw[icInk, line width=0.35pt] (-0.2,0.03) arc (180:360:0.2 and 0.07);
    \draw[icInk, line width=0.35pt] (-0.2,-0.09) arc (180:360:0.2 and 0.07);},
  ic-scorer/.pic={
    \draw[icInk, line width=0.45pt] (-0.27,-0.24) -- (0.27,-0.24);
    \foreach \x/\h in {-0.2/0.36,-0.08/0.22,0.04/0.42,0.16/0.28} \fill[icEval] (\x-0.045,-0.24) rectangle (\x+0.045,-0.24+\h);},
  ic-agent/.pic={
    \draw[icInk, fill=white, line width=0.45pt] (-0.26,-0.25) rectangle (0.12,0.25);
    \foreach \y in {0.13,0.0,-0.13} {\draw[icEval, line width=0.6pt] (-0.21,\y) -- (-0.18,\y-0.03) -- (-0.13,\y+0.04); \draw[black!40, line width=0.45pt] (-0.09,\y) -- (0.07,\y);}
    \draw[icMllm, fill=white, line width=0.7pt] (0.12,-0.04) circle (0.1);
    \draw[icMllm, line width=1.1pt] (0.19,-0.11) -- (0.27,-0.2);},
  ic-form/.pic={
    \draw[icInk, fill=white, line width=0.45pt, rounded corners=1pt] (-0.22,-0.27) rectangle (0.22,0.24);
    \fill[icInk] (-0.08,0.21) rectangle (0.08,0.28);
    \foreach \y/\c in {0.1/1,-0.04/0,-0.18/1} {\draw[icInk, line width=0.4pt] (-0.15,\y-0.04) rectangle (-0.07,\y+0.04);
      \ifnum\c=1 \draw[icEval, line width=0.6pt] (-0.15,\y) -- (-0.11,\y-0.04) -- (-0.06,\y+0.06); \fi
      \draw[black!40, line width=0.45pt] (-0.02,\y) -- (0.15,\y);}},
  ic-eye/.pic={
    \draw[icInk, fill=white, line width=0.55pt] (-0.27,0) .. controls (-0.1,0.18) and (0.1,0.18) .. (0.27,0) .. controls (0.1,-0.18) and (-0.1,-0.18) .. (-0.27,0) -- cycle;
    \fill[icInk] (0,0) circle (0.075); \fill[white] (0.025,0.025) circle (0.02);},
  ic-eyeoff/.pic={
    \pic {ic-eye}; \draw[icRed, line width=0.9pt] (-0.22,-0.17) -- (0.22,0.17);},
  ic-camera/.pic={
    \draw[icInk, fill=black!8, line width=0.5pt, rounded corners=1pt] (-0.27,-0.16) rectangle (0.15,0.14);
    \draw[icInk, fill=black!25, line width=0.5pt] (0.15,0.0) -- (0.28,0.1) -- (0.28,-0.12) -- cycle;
    \draw[icInk, fill=white, line width=0.45pt] (-0.06,-0.01) circle (0.08);
    \fill[icInk] (-0.2,0.14) rectangle (-0.1,0.2);},
  ic-warn/.pic={
    \draw[icRed, fill=icRed!12, line width=0.6pt, line join=round] (0,0.25) -- (0.27,-0.22) -- (-0.27,-0.22) -- cycle;
    \fill[icRed] (-0.025,-0.06) rectangle (0.025,0.12); \fill[icRed] (0,-0.13) circle (0.03);},
  ic-sentence/.pic={
    \draw[icCond, fill=icCond!8, line width=0.5pt, rounded corners=1pt] (-0.28,-0.12) rectangle (0.28,0.12);
    \draw[icCond, line width=0.9pt] (-0.2,0.03) -- (0.2,0.03); \draw[icCond, line width=0.9pt] (-0.2,-0.04) -- (0.06,-0.04);},
  ic-timeline/.pic={
    \draw[icInk, line width=0.5pt, -{Stealth[length=2.2pt]}] (-0.28,-0.12) -- (0.29,-0.12);
    \foreach \x in {-0.2,-0.07,0.06,0.19} {\draw[icInk, line width=0.4pt] (\x,-0.15) -- (\x,-0.09); \fill[icCond!75] (\x-0.04,0.0) rectangle (\x+0.04,0.17);}},
  ic-graph/.pic={
    \draw[icInk, line width=0.5pt] (-0.18,0.12) -- (0.17,0.15) -- (0.05,-0.15) -- cycle;
    \draw[icRed, line width=0.7pt, dashed, dash pattern=on 1.4pt off 0.9pt] (0.05,-0.15) -- (-0.22,-0.18);
    \foreach \p/\c in {(-0.18,0.12)/icCond,(0.17,0.15)/icCond,(0.05,-0.15)/icCond,(-0.22,-0.18)/icRed} \fill[\c] \p circle (0.055);},
  ic-chunks/.pic={
    \foreach \x in {-0.27,-0.09,0.09} {\draw[icInk, fill=icCond!10, line width=0.4pt] (\x,-0.2) rectangle (\x+0.17,0.2);
      \draw[icCond, line width=0.6pt] (\x+0.03,0.09) -- (\x+0.14,0.09) (\x+0.03,0.01) -- (\x+0.12,0.01) (\x+0.03,-0.07) -- (\x+0.1,-0.07);}},
  ic-chunksg/.pic={
    \pic {ic-chunks}; \foreach \x in {-0.185,-0.005,0.175} {\fill[icRed] (\x-0.03,-0.15) circle (0.022); \fill[icRed] (\x+0.03,-0.15) circle (0.022); \draw[icRed, line width=0.35pt] (\x-0.03,-0.15) -- (\x+0.03,-0.15);}},
  ic-narrative/.pic={
    \draw[icCond, fill=icCond!8, line width=0.5pt, rounded corners=1pt] (-0.25,-0.22) rectangle (0.25,0.22);
    \foreach \y/\l in {0.12/0.36,0.04/0.4,-0.04/0.33,-0.12/0.38} \draw[icCond, line width=0.6pt] (-0.19,\y) -- ++(\l,0);},
}

\tikzset{inl/.style={text width=, minimum width=0pt, minimum height=0pt, align=center, draw=none, fill=none, rounded corners=0pt}}

\newsavebox{\tframebox}
\newcommand{\tframe}[4][]{{%
  \sbox{\tframebox}{\includegraphics[width=#2]{#3}}%
  \if\relax\detokenize{#1}\relax\else
    \sbox{\tframebox}{\setlength{\fboxsep}{0pt}\setlength{\fboxrule}{0.9pt}\fcolorbox{#1}{white}{\usebox{\tframebox}}}\fi
  \usebox{\tframebox}%
  \if\relax\detokenize{#4}\relax\else
    \hspace{-\wd\tframebox}\makebox[\wd\tframebox][l]{\raisebox{\dimexpr\ht\tframebox-5.4pt\relax}[0pt][0pt]{%
      \setlength{\fboxsep}{0.6pt}\colorbox{white}{\fontsize{4.6}{5}\selectfont #4}}}\fi}}
\definecolor{fMis}{HTML}{D62728}\definecolor{fL1}{HTML}{2F6DB5}\definecolor{fL2}{HTML}{B86E00}
\definecolor{fL3}{HTML}{2E8B57}\definecolor{fAnn}{HTML}{7A4FB3}\definecolor{fTpl}{HTML}{77766F}

\graphicspath{{figures/}}

\newif\iftrimpreview
\trimpreviewfalse

\newif\ifrevhl
\revhlfalse
\DeclareRobustCommand{\rev}[1]{\ifrevhl\leavevmode{\color{blue}#1}\else #1\fi}

\definecolor{bestcell}{HTML}{C8E6C9}
\definecolor{secondcell}{HTML}{FFE0B2}
\definecolor{tableheader}{HTML}{F1F5F9}
\definecolor{subsubheader}{HTML}{F8FAFC}
\newcommand{\bestmark}[1]{\cellcolor{bestcell}\textbf{#1}}
\newcommand{\secondmark}[1]{\cellcolor{secondcell}\textbf{#1}}

\DeclareFontFamily{U}{dutchcal}{\skewchar\font=45}
\DeclareFontShape{U}{dutchcal}{m}{n}{<-> dutchcal-r}{}
\DeclareFontShape{U}{dutchcal}{b}{n}{<-> dutchcal-b}{}
\DeclareFontShape{U}{dutchcal}{bx}{n}{<-> ssub * dutchcal/b/n}{}

\newcommand{\vewomood}{\textbf{\textsc{VeWomOOD}}}
\newcommand{\benchname}{\vewomood{} \textbf{\textsc{Dataset}}}                
\newcommand{\vewomooda}{\texorpdfstring{\vewomood{}$_{\mathbf{A}}$}{VeWomOOD-A}} 
\newcommand{\evalname}{\textbf{\textsc{VeWomEval}}}
\newcommand{\evalframework}{\evalname{} \textbf{\textsc{Framework}}}         
\newcommand{\evalprotocol}{\evalname{} \textbf{\textsc{Protocol}}}            
\newcommand{\evaldataset}{\evalname{} \textbf{\textsc{Dataset}}}              

\newcommand{\VDMFull}{Video Generation Models}

\newcommand{\VDMs}{VGMs}
\newcommand{\VDM}{VGM}

\newcommand{\tediopp}{TeDiO++}
\newcommand{\bachvidpp}{BachVid++}
\newcommand{\ttoname}{MemAlignTTO}                          
\newcommand{\steeraxis}{Denoising Trajectory Steering}
\newcommand{\steeraxislow}{denoising trajectory steering}

\newcommand{\steeraxisfulllow}{\steeraxislow{} axis}
\newcommand{\promptaxis}{Prompt Conditioning}
\newcommand{\promptaxislow}{prompt conditioning}

\newcommand{\promptaxisfulllow}{\promptaxislow{} axis}

\usepackage{listings}
\usepackage{etoolbox}
\usepackage{etoc}
\makeatletter
\patchcmd{\LT@output}{\vss}{\vskip\z@\@plus 1fil\@minus\vsize}{}{}
\patchcmd{\LT@output}{\vss}{\vskip\z@\@plus 1fil\@minus\vsize}{}{}
\AtBeginDocument{\immediate\write\@auxout
  {\string\@writefile{toc}{\string\etoc@depthtag{mainpaper}}}}
\newcommand{\supplementarycontents}{%
  \def\cvprsect##1{\cvprsection[##1]{\texorpdfstring{\hskip -1em.~}{}##1}}%
  \def\cvprsubsect##1{\cvprsubsection[##1]{\texorpdfstring{\hskip -1em.~}{}##1}}%
  \def\cvprsubsubsect##1{\cvprsubsubsection[##1]{\texorpdfstring{\hskip -1em.~}{}##1}}%
  \etocdepthtag.toc{supplementary}%
  \etocsettagdepth{mainpaper}{none}%
  \etocsettagdepth{supplementary}{subsubsection}%
  \begingroup
  \patchcmd{\l@section}{1.0em}{0.5em}{}{}%
  \renewcommand{\@pnumwidth}{2.3em}
  \renewcommand{\@tocrmarg}{3.3em}%
  \tableofcontents
  \endgroup}
\newcommand{\appheading}{\@startsection{paragraph}{4}{\z@}
  {2.2ex plus 0.6ex minus 0.2ex}{0.8ex}{\normalsize\bfseries\itshape}*}
\newcommand{\unlinkedexternaldocument}[1]{%
  \let\orig@XR@@dURL\XR@@dURL
  \def\XR@@dURL##1##2##3##4##5\\{\unexpanded{{##1}{##2}{##3}{}{}}}%
  \externaldocument[][nocite]{#1}[]%
  \let\XR@@dURL\orig@XR@@dURL}
\makeatother

\usepackage[below]{placeins}
\extrafloats{100}

\makeatletter
\patchcmd{\lst@frameInit}{\noindent\box\z@\par}%
  {\begingroup\@ifundefined{nolinenumbers}{}{\nolinenumbers}\noindent\box\z@\par\endgroup}{}{}
\patchcmd{\lst@frameExit}{\everypar{}\par\lst@parshape\nointerlineskip\noindent\box\z@}%
  {\begingroup\@ifundefined{nolinenumbers}{}{\nolinenumbers}\everypar{}\par\lst@parshape\nointerlineskip\noindent\box\z@\par\endgroup}{}{}
\makeatother

\usepackage{xr-hyper}

\definecolor{cvprblue}{rgb}{0.21,0.49,0.74}
\usepackage[pagebackref,breaklinks,colorlinks,allcolors=cvprblue]{hyperref}

\IfFileExists{supp.aux}{\unlinkedexternaldocument{supp}}{\input{paper_tex_files/supp_labels}}

\def\confName{CVPR}
\def\confYear{2027}

\title{Dreaming in Distribution: Understanding the Out-of-Distribution\\Capabilities of \VDMFull{}}

\author{
  Rohith Peddi$^{1}$ \qquad Shravan Shanmugam$^{1}$ \qquad Likhitha Pallapothula$^{1}$ \\
  Ayush Shrivastava$^{2}$ \qquad Parag Singla$^{3}$ \qquad Vibhav Gogate$^{1}$ \\[1.2ex]
  $^{1}$UT Dallas \qquad $^{2}$UCLA \qquad $^{3}$IIT Delhi \\[1.0ex]
  {\small\textbf{Codebase:} \url{https://github.com/rohithpeddi/VeWomRelease}}
}

\begin{document}
\maketitle
\vspace{-2mm}
\begin{abstract}
\VDMFull{} (\VDMs{}) synthesize visually plausible motion, yet it remains unclear whether they can accurately continue an observed scene when impending physical cues point toward an unintentional action rather than a routine activity. To this end, we introduce the \benchname{}, comprising 1,487 Point of No Return (PNR) frame-annotated videos of real-world unintentional actions. Leveraging this dataset, we systematically evaluate frozen \VDMs{} along two inference-time axes: (1) \emph{\steeraxis{}}, in which our proposed training-free module combines enhanced versions of two existing inference-time video generation methods with a novel cross-chunk memory component, \ttoname{}; and (2) \emph{\promptaxis{}}, in which an agentic planning-and-verification framework infers hierarchical MLLM semantic guidance from pre-PNR context. Furthermore, we propose an evaluation framework called \evalname{} that includes (a) a human-annotated dataset (\evaldataset{}) for cross-verification with MLLM judges, and (b) a unified tool and skill design (\evalprotocol{}) for an MLLM judge to select perception evidence and understand generated videos clearly. We make an interesting observation that \VDMs{} without semantic guidance cleverly alter scene dynamics toward routine actions, whereas explicit pre-PNR semantic guidance steers continuations closer to the recorded unintentional actions.
\end{abstract}

\par\vspace{-2mm}

\par\vspace{-2mm}
\section{Introduction}
\label{sec:intro}
\vspace{-2mm}
\begin{figure}[t]
\centering
\includegraphics[width=\columnwidth]{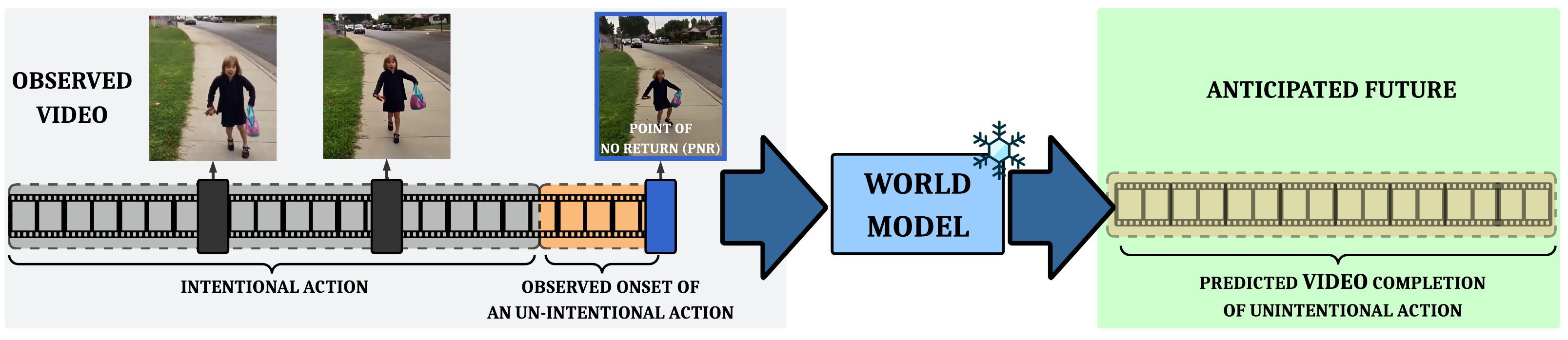}
\vspace{-8mm}
\caption{\textbf{Overview.} \textbf{Left:} The input video begins with a routine \emph{Intentional Action} (gray segment, where callout frames show a child running along a sidewalk) followed by the \emph{Observed Onset of an Unintentional Action} (orange segment), terminating at the human-annotated \emph{Point of No Return} (\textbf{PNR}, blue marker, where she begins stumbling). \textbf{Center:} A frozen world model, here a \VDM{}, observes only the pre-PNR video. \textbf{Right:} The model synthesizes the \emph{Anticipated Future}, i.e., the predicted video completion of the unintentional action after the PNR.}
\label{fig:overview}
\vspace{-6mm}
\end{figure}

\begin{figure*}[t]
\centering
\includegraphics[width=\linewidth]{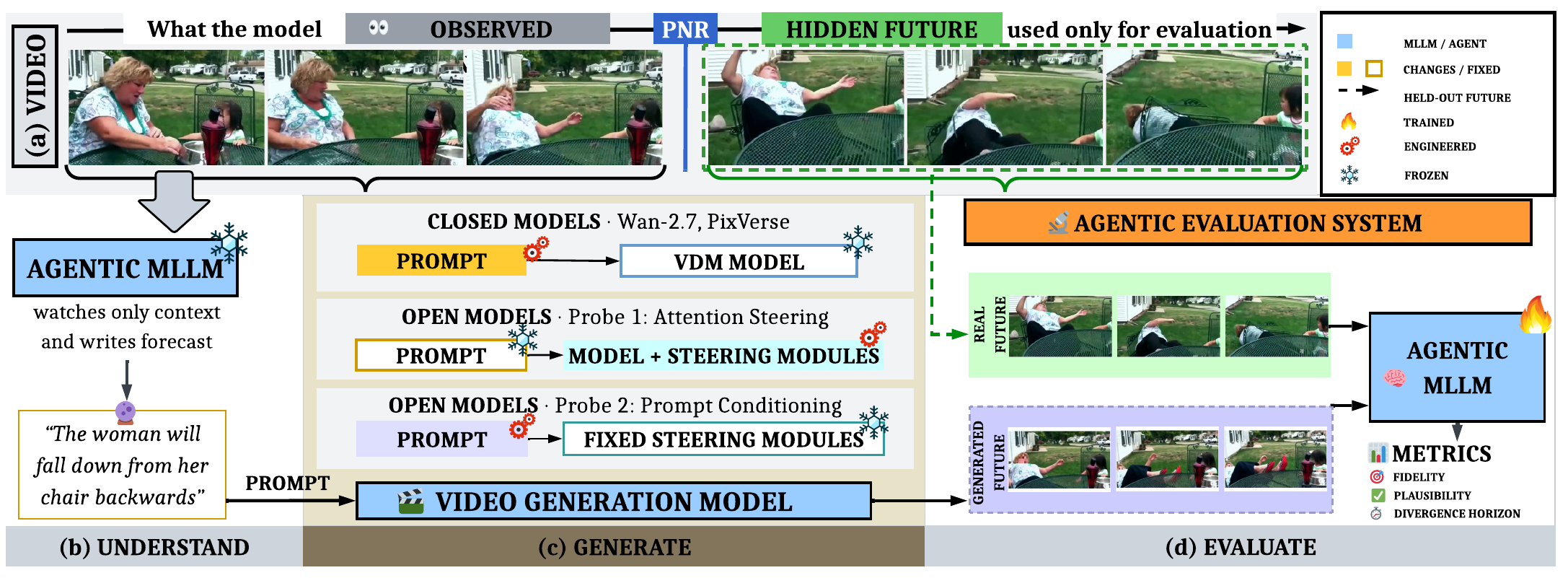}
\vspace{-8mm}
\caption{\textbf{Overview of Task Setup.} \underline{\textbf{(b) Understand:}} Given pre-PNR context as shown in \textbf{(a) Video}, an agentic \emph{MLLM} inspects only the pre-PNR context to infer verified hierarchical semantic guidance $A(\mathcal{C})$. \underline{\textbf{(c) Generate:}} Compiled prompts $\Phi_\ell(A(\mathcal{C}))$ condition a frozen \emph{\VDM{}} ($G_\theta$) across three settings: \emph{Closed Models}; \emph{Open Models, Probe~1} (\steeraxislow{}); and \emph{Open Models, Probe~2} (\promptaxislow{}). \underline{\textbf{(d) Evaluate:}} The \emph{Agentic Evaluation System} then compares the \emph{Real Future} and \emph{Generated Future}.}
\label{fig:task}
\vspace{-6mm}
\end{figure*}

Modern \VDMFull{} (\VDMs{}) synthesize high-fidelity visual motion, motivating growing interest in employing them as general-purpose world models \citep{ho2022video,brooks2024sora,wan2025wan}. However, visual realism alone does not guarantee that a generator can follow physical cues toward an unexpected, out-of-distribution (OOD) outcome. In this work, we investigate this gap between \emph{synthesizing a visually plausible continuation} and \emph{predicting an unexpected real-world outcome} (Figure~\ref{fig:overview}). To tackle this challenge, we curate the \benchname{} from real-world unintentional actions across CaptainCook4D \citep{peddi2024captaincook4d} and OOPS \citep{epstein2020oops}. Every video segment is split at a human-annotated Point of No Return (PNR), separating the observed visual context from the held-out continuation.\footnote{Rather than claiming strict statistical disjointness from undisclosed pretraining corpora, we employ the term \emph{out-of-distribution} (OOD) operationally to characterize departures from routine, everyday execution.}


We organize our study around two complementary inference-time axes (Figure~\ref{fig:task}): (1) \textbf{\steeraxis{}} modifies intermediate attention outputs and noisy latent states inside open-weight \VDMs{} during denoising through our proposed training-free \rev{Combined Steering} module, which encapsulates enhanced continuation versions of two previously proposed methods (\tediopp{} \citep{tursynbek2026tedio} and \bachvidpp{} \citep{bachvid2025}) in combination with a novel retrieval-gated memory component (\ttoname{}). (2) \textbf{\promptaxis{}} varies the hierarchical MLLM semantic guidance supplied in the text prompt, progressing from a prediction-free \emph{baseline} control to coarse unintentional-action alerts, event summaries, interaction summaries, and unlocalized scene-graph summaries.




A generated clip can look physically realistic while depicting the wrong outcome, or depict the matching outcome at a different speed than the reference recording. To address this, we propose the \evalframework{}, which unifies two components. First, the \evalprotocol{} equips an agentic evaluator with vision foundation models as perception tools and deterministic physics skills. Together, they keep event fidelity, physical plausibility, visual quality, and prompt adherence strictly separate. Second, to correlate human judgments with agentic MLLM judgment capabilities, we use the \evaldataset{}, in which human reviewers narrate anomalies and divergences in generated videos. Thus, this paper makes the following contributions:
\begin{itemize}
    \item[--] \benchname{}\textbf{.} We curate 1,487 PNR-annotated videos of real-world unintentional actions.
    \item[--] \textbf{Generation Framework.} We formulate a unified inference-time framework across \VDMs{} that pairs (i) a training-free \steeraxislow{} module combining \tediopp{}, \bachvidpp{}, and our novel component \ttoname{} with (ii) an agentic planning-and-verification pipeline that infers semantic guidance.
    \item[--] \evalframework{}\textbf{.} We introduce the \evalprotocol{}, combining perception tools, physics skills, an agentic evaluator, and a 12-axis perception scorer, together with the human-annotated \evaldataset{} for cross-verifying MLLM judges.
\end{itemize}

\par\vspace{-2mm}

\vspace{-2mm}
\section{Related Work}
\label{sec:related}
\vspace{-2mm}
\noindent\textbf{World Models.} Conditional \VDMs{} predict future frames from observed visual history \citep{voleti2022mcvd,harvey2022flexible,zhang2024extdm}. This capability has motivated recent efforts to treat large-scale video generators as implicit physical world models\footnote{To the best of our knowledge, this represents the first study on video extension for generating videos corresponding to unintentional actions.} \citep{brooks2024sora,wan2025wan,hacohen2024ltxvideo}. Language-guided pipelines connect high-level planning with pixel synthesis by conditioning video backbones on predicted future states \citep{xing2025aid,chi2025empowering,cheng2026vans}.

\noindent\textbf{Inference interventions.} Recent test-time methods extend generation length and improve temporal consistency without retraining by modifying noise schedules, attention windows, or rotary positional encodings \citep{qiu2024freenoise,lu2024freelong,zhao2025riflex}. Related latent-guidance and particle-sampling techniques steer trajectory semantics and sample diversity \citep{chefer2023attendexcite,epstein2023selfguidance,corso2024particle,sadat2024cads}\rev{, complementing robust probabilistic formulations that guard against out-of-distribution shifts \citep{peddi2022robust,peddi2022distributionally}}.


\noindent\textbf{Anticipating Unintentional Actions.} Classical action and scene-graph anticipation methods predict upcoming activities or object relations in symbolic or text space \citep{furnari2019rulstm,grauman2022ego4d,peddi2024scenesayer,peddi2025impartail}\rev{, while recent world-centric scene-graph formulations maintain persistent 3D object relations across occlusions \citep{peddi2026worldsgg}}. Meanwhile, benchmarks such as CaptainCook4D \citep{peddi2024captaincook4d}, OOPS \citep{epstein2020oops}, \rev{and Mistake Attribution \citep{li2026mistake}} curate real-world unintentional actions\rev{, motivating proactive augmented-reality task guidance \citep{rheault2024predictive}}. 


\noindent\textbf{Evaluating Video Generation.} Standard video generation metrics such as FVD \citep{unterthiner2018fvd} and VBench \citep{huang2024vbench} measure distributional similarity and visual quality. More recent benchmarks, including VideoPhy \citep{bansal2024videophy}, \rev{VideoPhy-2 \citep{bansal2025videophy2},} WorldModelBench \citep{li2025worldmodelbench}, \rev{PhyWorldBench \citep{gu2026phyworldbench}, CausalCascade \citep{li2026causalcascade}, Physion-Eval \citep{zhang2026physioneval}, Foresight at the Event Boundary \citep{santana2026foresight},} and Physics-IQ \citep{motamed2025physicsiq}, test physical commonsense and short-horizon dynamics. Evaluating \vewomood{} continuations requires going beyond visual quality or pixel overlap. Specifically, we must determine \emph{which} event unfolded, \emph{which} entities participated, and whether physical laws were held true even when a generated clip diverges from the ground-truth future (see App.~\ref{app:related} for related work).

\par\vspace{-2mm}

\begin{figure*}[t]
\centering
\includegraphics[width=\linewidth]{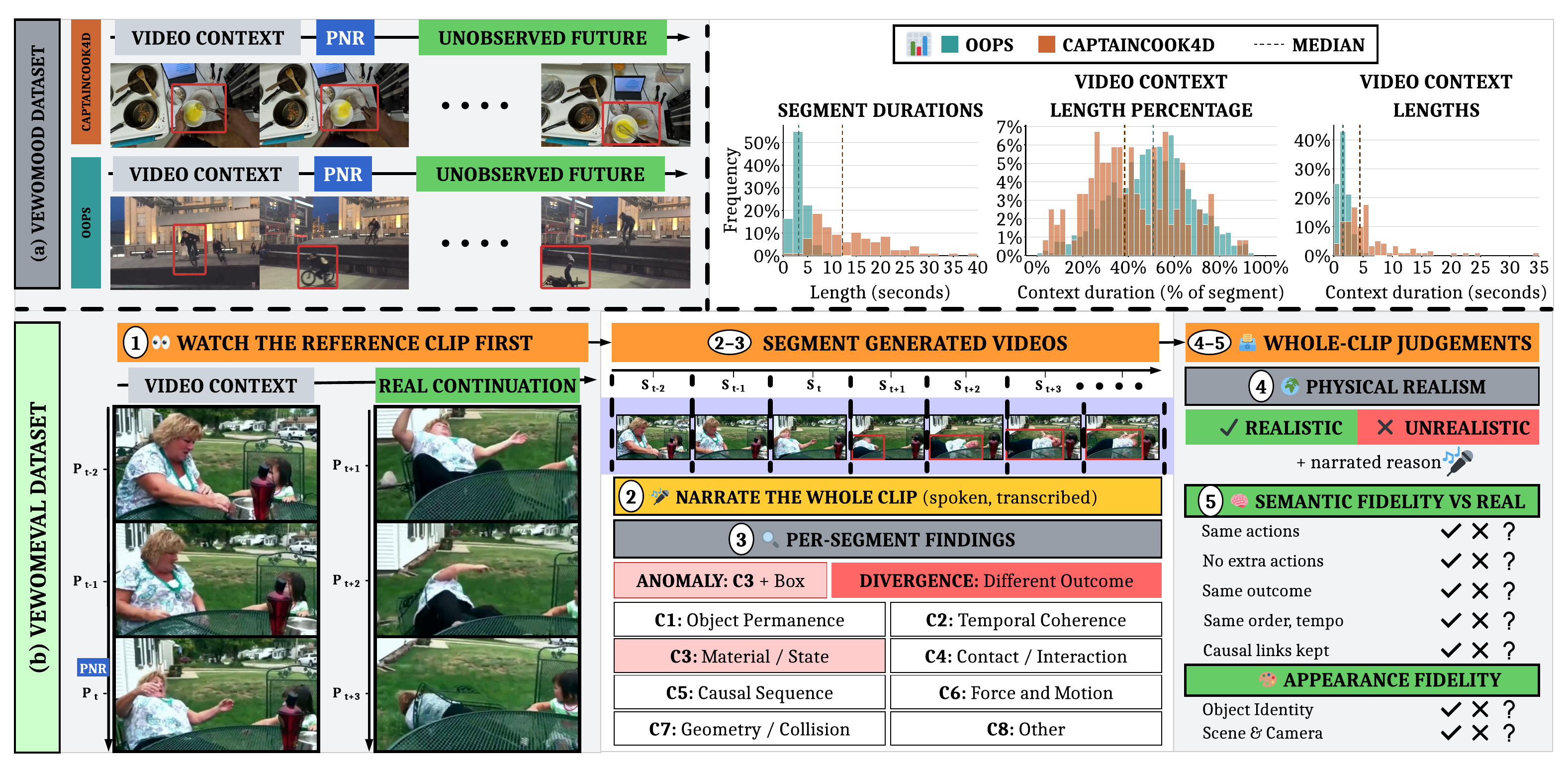}
\vspace{-8mm}
\caption{\textbf{Overview of the \benchname{} and the \evaldataset{}.} \textbf{(a) \benchname{} (top):} Representative clips from \emph{OOPS} and \emph{CaptainCook4D} progress from \emph{Video Context} across the \emph{PNR} into the \emph{Unobserved Future}. \textbf{(b) \evaldataset{} (bottom, steps 1--5):} Summary of all human-annotated fields provided by the dataset for generated video continuations.}
\label{fig:dataset}
\vspace{-2mm}
\end{figure*}

\vspace{-2mm}
\section{Notation and Problem Description}
\label{sec:task_definition}
\label{sec:notation}
\vspace{-2mm}
Let $V=(x_1,\ldots,x_T)$ denote a real-world video segment comprising $T$ frames bounded by a decision window $[s_{\mathrm{start}}, s_{\mathrm{end}}]$, where visible physical or procedural cues precede an unintentional action (Figures~\ref{fig:overview} and~\ref{fig:task}a). We summarize all symbols in Table~\ref{tab:notation} of App.~\ref{app:notation}. Let $s_{\mathrm{pnr}} \in (s_{\mathrm{start}}, s_{\mathrm{end}})$ denote the human-annotated Point of No Return (PNR) timestamp at which the unintentional action becomes unavoidable, with corresponding frame index $t_{\mathrm{pnr}}$ in $V$. Thus, the PNR splits $V$ into two disjoint sequences: the observed pre-PNR visual context $\mathcal{C}=(x_1,\ldots,x_{t_{\mathrm{pnr}}})$ and the held-out ground-truth continuation $Y^*=(x_{t_{\mathrm{pnr}}+1},\ldots,x_T)$. Because source recordings and video generators run at different native frame rates, we index post-PNR states on a shared clock measured in seconds after $s_{\mathrm{pnr}}$ via $\tau_i = i / f$ for $i \in \{1,\ldots,K\}$, where $f = 3$\,Hz denotes the prediction rate and $K$ denotes the number of discrete post-PNR steps. Let $A$ denote our agentic anticipation pipeline and $G^{\mathrm{obs}}$ denote the scene graph observed at $t_{\mathrm{pnr}}$. Observing strictly $\mathcal{C}$, $A$ infers a verified four-part semantic guidance hierarchy $A(\mathcal{C}) = (e, \mathbf{L1}, \mathbf{L2}, \mathbf{L3})$. Specifically, $A(\mathcal{C})$ comprises an early unintentional-action alert $e$, a global event summary \textbf{L1}, timestamped interaction summaries \textbf{L2} at $\{\tau_i\}_{i=1}^K$, and unlocalized scene-graph summaries \textbf{L3} initialized from $G^{\mathrm{obs}}$. We then employ a deterministic prompt compiler $\Phi_\ell$ to map subsets of $A(\mathcal{C})$ into structured conditioning text $c = \Phi_\ell(A(\mathcal{C}))$ for prompt arm $\ell \in \{1,\ldots,6\}$. Finally, we denote a frozen \VDM{} with fixed weights $\theta$ by $G_\theta$, parameterized at test time by steering configuration $P$ and random seed $\xi$, where each call evolves $L$ latent frames across denoising steps $\zeta$ and updates the generated noisy latent slice $z_{\mathrm{gen}}$.

Given strictly $\mathcal{C}$ while withholding $Y^*$ for evaluation, we test whether $G_\theta$ can synthesize a continuation $\hat{Y}$ past $t_{\mathrm{pnr}}$ that captures the impending unintentional action rather than reverting to routine activity (Figure~\ref{fig:task}). Formally, the generated continuation $\hat{Y}_{P,\ell}$ under steering configuration $P$ and prompt arm $\ell$ is given by
\begin{equation}
\hat{Y}_{P,\ell}=G_\theta\bigl(\mathcal{C},\Phi_\ell(A(\mathcal{C}));P,\xi\bigr)
\label{eq:probe}
\end{equation}
Here, we hold $\mathcal{C}$, the target duration, and $\xi$ fixed across compared arms. Along the \textbf{\steeraxisfulllow{}} (Probe~1), we fix $\Phi_\ell$ and vary $P$ to steer intermediate DiT attention features and $z_{\mathrm{gen}}$. Conversely, along the \textbf{\promptaxisfulllow{}} (Probe~2), we fix $P$ and vary $\Phi_\ell$. When the target horizon exceeds one call, we unroll $G_\theta$ across chunks $m$ over post-PNR windows $W_m$ with chunk-aligned prompts $c_{\mathrm{txt}}^{(m)} = \Phi_{\ell,m}(A(\mathcal{C}))$, reusing trailing generated frames as rolling context.



\vspace{-1mm}
\section{\vewomood{} and \evalname{} Datasets}
\label{sec:benchmark}
\vspace{-2mm}

\label{sec:vewomood_dataset}
\label{sec:data_sources}
\label{sec:pnr_stats}
\noindent\textbf{VeWomOOD Dataset.}
We curate the \benchname{} from two complementary sources of real-world unintentional actions: egocentric cooking errors in CaptainCook4D \citep{peddi2024captaincook4d} and third-person unintentional actions in OOPS \citep{epstein2020oops}. Starting from \rev{30,147 candidate recordings yielding} 1,561 sampled OOPS videos and 148 CaptainCook4D segments bounded by step- or action-level windows $[s_{\mathrm{start}}, s_{\mathrm{end}}]$, we employ a frame-stepping web interface (\rev{Fig.~\ref{fig:pnr-tool},} App.~\ref{app:dataset}) to annotate $s_{\mathrm{pnr}}$ and split each clip into $\mathcal{C}$ and $Y^*$. We then filter out blurry footage, out-of-frame focal objects, segments lasting $40$\,s or longer, and segments lacking complete multi-arm continuations. Thus, the \benchname{} comprises \textbf{1,487 segments}, spanning \textbf{1,373 segments from OOPS} (1,325 videos) and \textbf{114 segments from CaptainCook4D} (80 recordings across 22 recipes\rev{; 57 action types and 2,660 object categories, Tables~\ref{tab:dataset-stats}--\ref{tab:cohort-provenance}}). As shown in Figure~\ref{fig:dataset}a, OOPS captures abrupt short-horizon unintentional actions (median duration $3.15$\,s, median context $1.51$\,s, PNR at $51\%$), whereas CaptainCook4D captures extended procedural unintentional actions (median duration $12.6$\,s, median context $4.70$\,s, PNR at $39\%$) requiring multi-chunk rollouts.

To validate PNR boundary reliability, we conduct an inter-annotator study on $N_{\mathrm{pnr}} = 100$ double-annotated segments ($92$ OOPS and $8$ CaptainCook4D) labeled independently by $2$ annotators. Given independent annotations $(s_{\mathrm{pnr}}^{(1)}, t_{\mathrm{pnr}}^{(1)})$ and $(s_{\mathrm{pnr}}^{(2)}, t_{\mathrm{pnr}}^{(2)})$, we first compute the \textbf{Mean Absolute Temporal Offset} $|\Delta s_{\mathrm{pnr}}| = |s_{\mathrm{pnr}}^{(1)} - s_{\mathrm{pnr}}^{(2)}|$, which averages $0.03$\,s ($0.7$ frames). Second, we evaluate \textbf{Windowed Temporal IoU} $\mathrm{tIoU}_w = |I_w^{(1)} \cap I_w^{(2)}| / |I_w^{(1)} \cup I_w^{(2)}|$ over symmetric $(2w+1)$-frame intervals $I_w^{(r)} = [t_{\mathrm{pnr}}^{(r)} - w,\, t_{\mathrm{pnr}}^{(r)} + w]$, reaching $0.87$ at $w = 4$ ($9$-frame window, with $0.88$ on OOPS and $0.76$ on CaptainCook4D; see App.~\ref{app:dataset}).

\smallskip
\label{sec:evaldataset}
\noindent\textbf{VeWomEval Dataset.}
While we evaluate open-weight generators across all six prompt arms on the complete 1,487-segment \benchname{}, commercial API constraints motivate selecting a 510-segment cross-generator subset, denoted \vewomooda{} (455 OOPS and 55 CaptainCook4D segments; App.~\ref{app:conditioning}), with valid continuations across all four open- and closed-weight generators under \emph{zero-shot/baseline} and unintentional-action \emph{detection} prompts. Constructed on top of the generated continuations of \vewomooda{}, the \evaldataset{} provides strictly \emph{human annotations} across \textbf{4,096 videos} to diagnose continuation failure modes and calibrate automated evaluators (see App.~\ref{app:evalset}). Across ten equal post-PNR timeline bins, annotators provide (1) a \emph{whole-clip narration}, (2) localized \emph{physical anomalies} typed across eight categories (\texttt{C1}--\texttt{C8}) with severity ($\rev{1}$--$5$) and (3) localized \emph{outcome divergences}. At the whole-clip level, we collect (4) a \rev{tri-state} \emph{physical realism gate}. We also collect (5) tri-state (\rev{\emph{pass}/\emph{fail}/\emph{abstain}}) ratings across five \emph{semantic fidelity} axes (same actions \texttt{SA}, no extra actions \texttt{NH}, same outcome \texttt{SO}, sequence order and tempo \texttt{SQ}, and causal links \texttt{CL}) and two \emph{appearance fidelity} axes (object identity \texttt{OI} and scene/camera consistency \texttt{SC}), alongside a three-way \emph{event verdict} (\emph{same event}, \emph{different outcome}, or \emph{undecidable}).

Because our primary store maintains one shared human record per clip, we validate human--human reliability on an anonymized, independently double-annotated subset of $200$ clips across $64$ source segments rated by $2$ annotators without generator labels (\rev{App.~\ref{app:evalset}}). Specifically, we match post-PNR bin spans via Hungarian assignment at temporal $\mathrm{IoU} \ge 0.3$ to measure corpus-pooled $F_1^{\mathrm{anom}} = 0.78$ (with \texttt{C1}--\texttt{C8} category accuracy $86.4\%$) for physical anomalies and $F_1^{\mathrm{div}} = 0.81$ for outcome divergences.

\vspace{-1mm}
\section{Technical Approach: Generation}
\label{sec:vidgen}
\label{sec:vidgen_prelim}
\vspace{-2mm}

\begin{figure*}[!t]
\centering
\includegraphics[width=\linewidth]{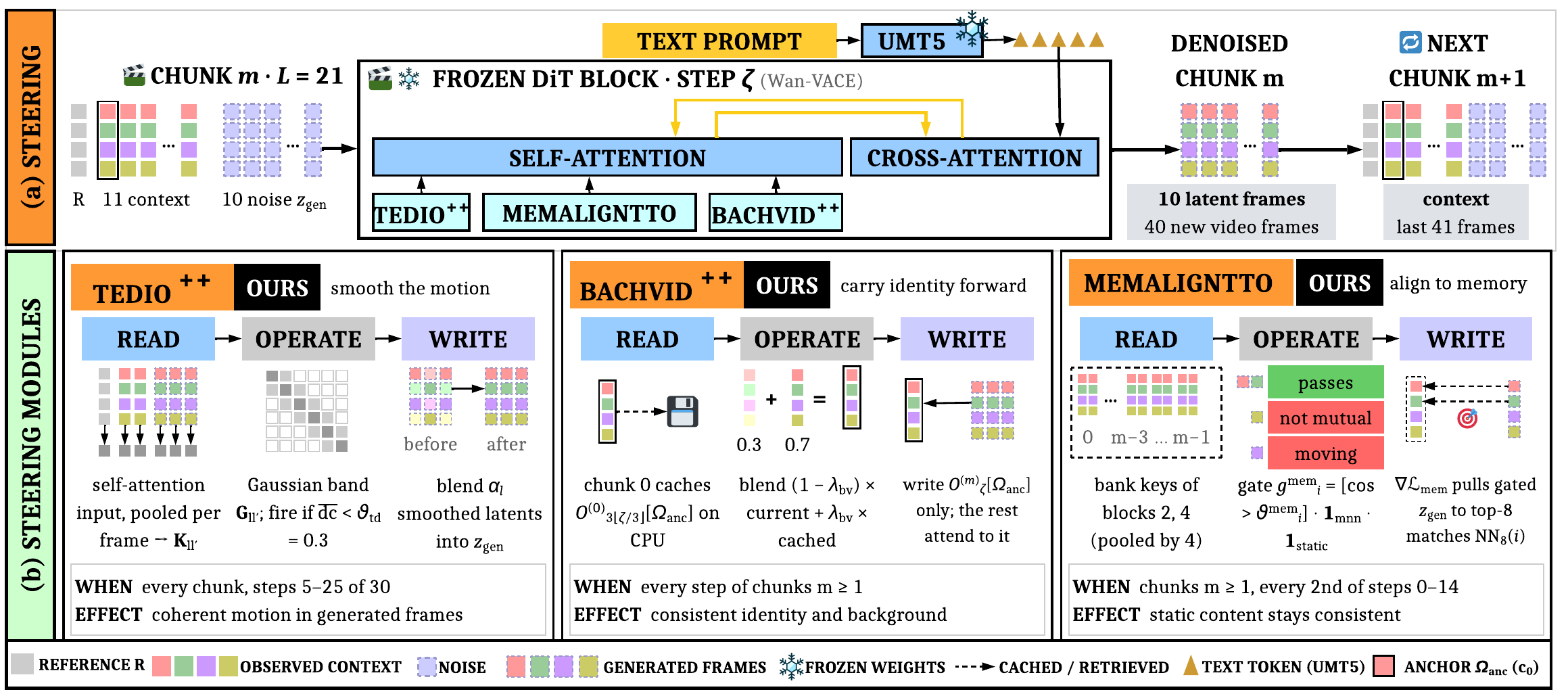}
\vspace{-8mm}
\caption{\textbf{\steeraxis{} Module.} \textbf{(a) Token Flow (top):} \textbf{Left:} The forward pass begins by combining a prepended reference frame \texttt{R}, rolling context frames, and noisy target frames. \textbf{Center:} Tokens traverse stacked frozen DiT blocks where \emph{Self-Attention} hosts our three steering hooks (\emph{\tediopp{}}, \emph{\ttoname{}}, and \emph{\bachvidpp{}}) while \emph{Cross-Attention} injects frozen \emph{UMT5} prompt tokens. \textbf{(b) Steering Modules (middle):} \textbf{Left (\tediopp{})} pools per-frame self-attention inputs, checks band-diagonal temporal concentration, and smooths noisy target frames across mid-schedule steps; \textbf{Center (\bachvidpp{})} caches first-chunk anchor outputs on CPU and blends them into the anchor slot of later chunks; and \textbf{Right (\ttoname{})} banks pooled shallow keys across chunks, filters static mutual-nearest-neighbor tokens, and pulls gated target latents toward retrieved bank matches.}
\label{fig:plugins}
\vspace{-6mm}
\end{figure*}

\subsection{\steeraxis{}}
\label{sec:vidgen_tto}

Using VACE models in the video extension setting, we observed that multi-chunk continuations degrade past the PNR due to three bottlenecks: (1) erratic frame-to-frame motion, (2) gradual drift in actor or background identity across chunks, and (3) loss of spatial correspondence once early frames exit the rolling window. To address these issues without retraining, we formulate a unified training-free \rev{Combined Steering} module (\texttt{TD+BV+MA}, Figure~\ref{fig:plugins}) over video DiT self-attention layers. Because masked V2V backbones constrain rolling context frames through a separate control branch, perturbing context latents directly creates visible seams at $t_{\mathrm{pnr}}$. Thus, we decouple feature reading from latent writing: we update only the generated noisy latent slice $z_{\mathrm{gen}}$ entering step $\zeta + 1$ in \tediopp{} and \ttoname{}, while blending context anchor features in self-attention output space in \bachvidpp{} (see App.~\ref{app:plugins}).

\noindent\underline{\textbf{Temporal motion smoothing (\tediopp{}).}}
To suppress erratic motion without materializing quadratic token-level temporal attention matrices \citep{tursynbek2026tedio}, we employ \tediopp{}. Specifically, we spatially pool self-attention inputs within each latent frame $l \in \{1,\dots,L\}$ to compute zero-clipped pairwise cosine similarities $\mathbf{K}_{ll'}$, weighted by a Gaussian band $\mathbf{G}_{ll'} = \exp\bigl(-(l - l')^2 / (2\sigma_{\mathrm{band}}^2)\bigr)$ of width $\sigma_{\mathrm{band}}$. Formally, the diagonal concentration score and conditional latent smoothing update are given by
\[
\begin{aligned}
\mathrm{dc} &= \frac{\sum_{l,l'} \mathbf{K}_{ll'}\,\mathbf{G}_{ll'}}{\sum_{l,l'} \mathbf{K}_{ll'}}, \\
z_{\mathrm{gen}} &\leftarrow (1 - \alpha_l)\,z_{\mathrm{gen}} + \alpha_l\,\bigl(\mathrm{gs} *_t z_{\mathrm{gen}}\bigr) \quad \text{if } \overline{\mathrm{dc}} < \vartheta_{\mathrm{td}}
\end{aligned}
\]
Here, $\overline{\mathrm{dc}}$ averages $\mathrm{dc}$ across DiT blocks and guidance branches, $\vartheta_{\mathrm{td}}$ is the trigger threshold, $\mathrm{gs} *_t$ denotes 1D temporal Gaussian convolution, and $\alpha_l$ is the blending weight applied solely to $z_{\mathrm{gen}}$.

\noindent\textbf{Anchor identity caching (\bachvidpp{}).}
To prevent actor and background drift after pre-PNR frames leave the rolling window, we adapt BachVid \citep{bachvid2025} into \bachvidpp{} within identity-sensitive DiT blocks. Specifically, during chunk $m = 0$, we cache self-attention outputs $O^{(0)}_\zeta[\Omega_{\mathrm{anc}}]$ at the initial context anchor span $\Omega_{\mathrm{anc}}$ every third denoising step and blend them into subsequent chunks $m \ge 1$:
\[
O^{(m)}_\zeta[\Omega_{\mathrm{anc}}] \leftarrow (1 - \lambda_{\mathrm{bv}})\,O^{(m)}_\zeta[\Omega_{\mathrm{anc}}] + \lambda_{\mathrm{bv}}\,O^{(0)}_{3\lfloor \zeta / 3 \rfloor}[\Omega_{\mathrm{anc}}]
\]
Here, $\lambda_{\mathrm{bv}}$ denotes the injection weight and $3\lfloor \zeta / 3 \rfloor$ indexes the latest cached step.


\noindent\underline{\textbf{Memory alignment (\ttoname{}).}}
Fixed anchor coordinates misalign under camera motion, yet naive cross-chunk attraction can pull an irreversible unintentional action back toward its pre-PNR state. Therefore, we propose \ttoname{} to replace fine-tuned memory queries \citep{memlearner2026} with training-free key retrieval. Specifically, we bank spatially pooled shallow self-attention keys across chunks and, at early denoising steps of chunks $m \ge 1$, compute cosine similarities $\cos_{ij}$ between generated probe keys $i$ and timestep-aligned bank keys $j$ ($\cos^{\max}_i = \max_j \cos_{ij}$). Formally, we gate each probe token $i$ by $g^{\mathrm{mem}}_i$ and minimize a soft top-8 retrieval loss $\mathcal{L}_{\mathrm{mem}}$ given by
\[
\begin{aligned}
g^{\mathrm{mem}}_i &= \mathrm{sigm}\!\left(\frac{\cos^{\max}_i - \vartheta^{\mathrm{mem}}_i}{0.03}\right) \cdot \mathbf{1}_{\mathrm{mnn}}(i) \cdot \mathbf{1}_{\mathrm{static}}(i), \\
\mathcal{L}_{\mathrm{mem}} &= \operatorname*{mean}_{\text{blocks}} \frac{\sum_i g^{\mathrm{mem}}_i \bigl(1 - \sum_{j \in \mathrm{NN}_8(i)} w_{ij}\cos_{ij}\bigr)}{\sum_i g^{\mathrm{mem}}_i + \epsilon_{\mathrm{mem}}}
\end{aligned}
\]
Here, $\mathrm{sigm}(\cdot)$ is the logistic sigmoid, $\vartheta^{\mathrm{mem}}_i$ increases linearly with matched chunk age so stale pre-PNR memories do not reverse an unfolding unintentional action, $\mathbf{1}_{\mathrm{mnn}}(i)$ enforces mutual nearest neighbors against repetitive textures, and $\mathbf{1}_{\mathrm{static}}(i)$ masks out high-motion tokens. In the loss, $w_{ij}$ denotes temperature-scaled softmax weights over the eight nearest bank keys $\mathrm{NN}_8(i)$, and $\epsilon_{\mathrm{mem}}$ stabilizes normalization. We apply an RMS-normalized gradient step to update solely $z_{\mathrm{gen}}$.

\begin{figure*}[!t]
\centering
\includegraphics[width=\linewidth]{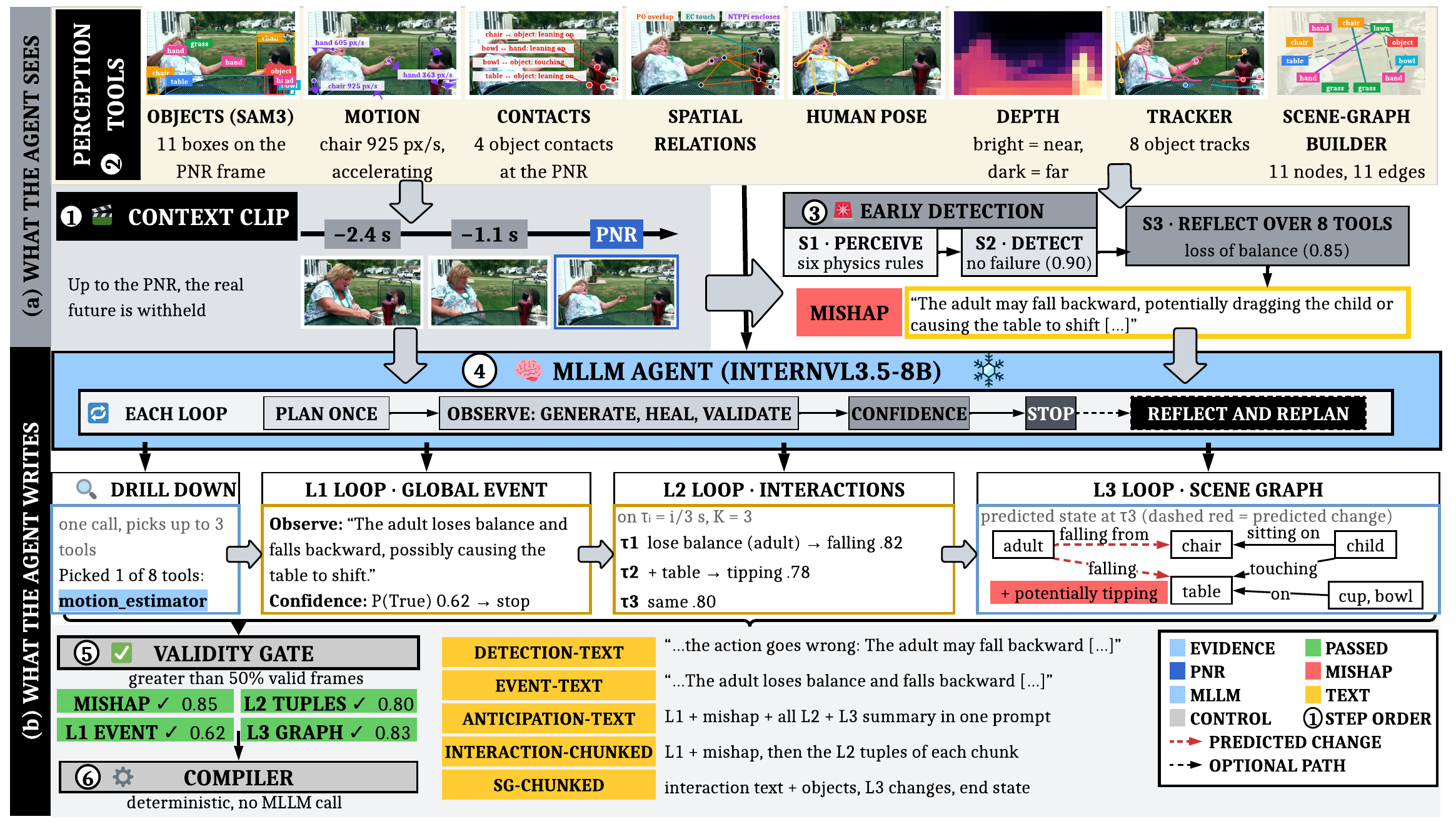}
\vspace{-8mm}
\caption{\textbf{Semantic guidance and Prompt compilation.} \textbf{(a) What the Agent Sees (top):} \textbf{(1) Top-left:} The agent observes only the $3.0$\,s pre-PNR \emph{Context Clip} ($\mathcal{C}$), while the ground-truth future ($Y^*$) is withheld. \textbf{(2) Bottom-left:} \emph{Early Detection} executes \emph{S1 Perceive}, \emph{S2 Detect}, and \emph{S3 Reflect Over 8 Tools}, emitting the \emph{Unintentional Action} alert ($e$). \textbf{(b) What the Agent Writes (bottom):} \textbf{(4) Top:} The \emph{MLLM Agent} executes coarse-to-fine planning, observation, confidence verification, and reflection/replanning across the \emph{L1 Loop}, \emph{L2 Loop}, and \emph{L3 Loop}. \textbf{(6) Bottom-right:} The deterministic \emph{Compiler} ($\Phi_\ell$) formats selected levels of $A(\mathcal{C})$ into six prompt arms.}
\label{fig:anticipation}
\vspace{-6mm}
\end{figure*}

\subsection{\promptaxis{}}
\label{sec:vidgen_antic}
\vspace{-2mm}
\noindent\textbf{Semantic guidance hierarchy.}
As introduced in Sec.~\ref{sec:task_definition} and Figure~\ref{fig:anticipation}, $A(\mathcal{C})$ structures pre-PNR guidance across four progressive levels of specificity. Specifically, (a) the \textbf{early unintentional-action alert} ($e$) states whether an unintended failure is imminent, which physical or procedural expectation fails, and its projected consequence; (b) the \textbf{global event summary (L1)} describes \emph{what} unintentional action unfolds across the post-PNR horizon; (c) the \textbf{interaction summary (L2)} specifies \emph{which} entities act or change state at each $3$\,Hz timestamp $\tau_i$ via concurrent action--object--tool--destination--state tuples; and (d) the \textbf{unlocalized scene-graph summary (L3)} records \emph{how} inter-object relations evolve from $G^{\mathrm{obs}}$ via added, removed, and modified subject--predicate--object edges.


\noindent\underline{\textbf{Agentic planning and verification.}}
To infer $A(\mathcal{C})$ strictly from $\mathcal{C}$ without ground-truth leakage, our two-stage agentic anticipation pipeline (Figure~\ref{fig:anticipation}) gives an MLLM access to vision foundation models as pre-PNR perception tools and leverages their outputs in its decision-making process. In Stage~1 (\emph{early detection}), the MLLM identifies actors, objects, and safety expectation rules ($\Psi$), inspects $\mathcal{C}$ for failure cues, and queries eight pre-PNR vision tools to emit $e$. In Stage~2 (\emph{hierarchical forecasting}), the agent issues a targeted tool drill-down query and executes a coarse-to-fine \textbf{Plan--Observe--Reflect--Replan} loop sequentially across \textbf{L1}, \textbf{L2}, and \textbf{L3}. App.~\ref{app:mllms} audits upstream forecast accuracy \rev{across seven MLLMs} against $Y^*$ \rev{(dual Gemini~3.1~Pro and Qwen3-VL-Plus references)} at $\{\tau_i\}_{i=1}^K$ relative to \emph{single-pass} and \emph{cascaded} baselines.



\begin{figure*}[t]
\centering
\includegraphics[width=\linewidth]{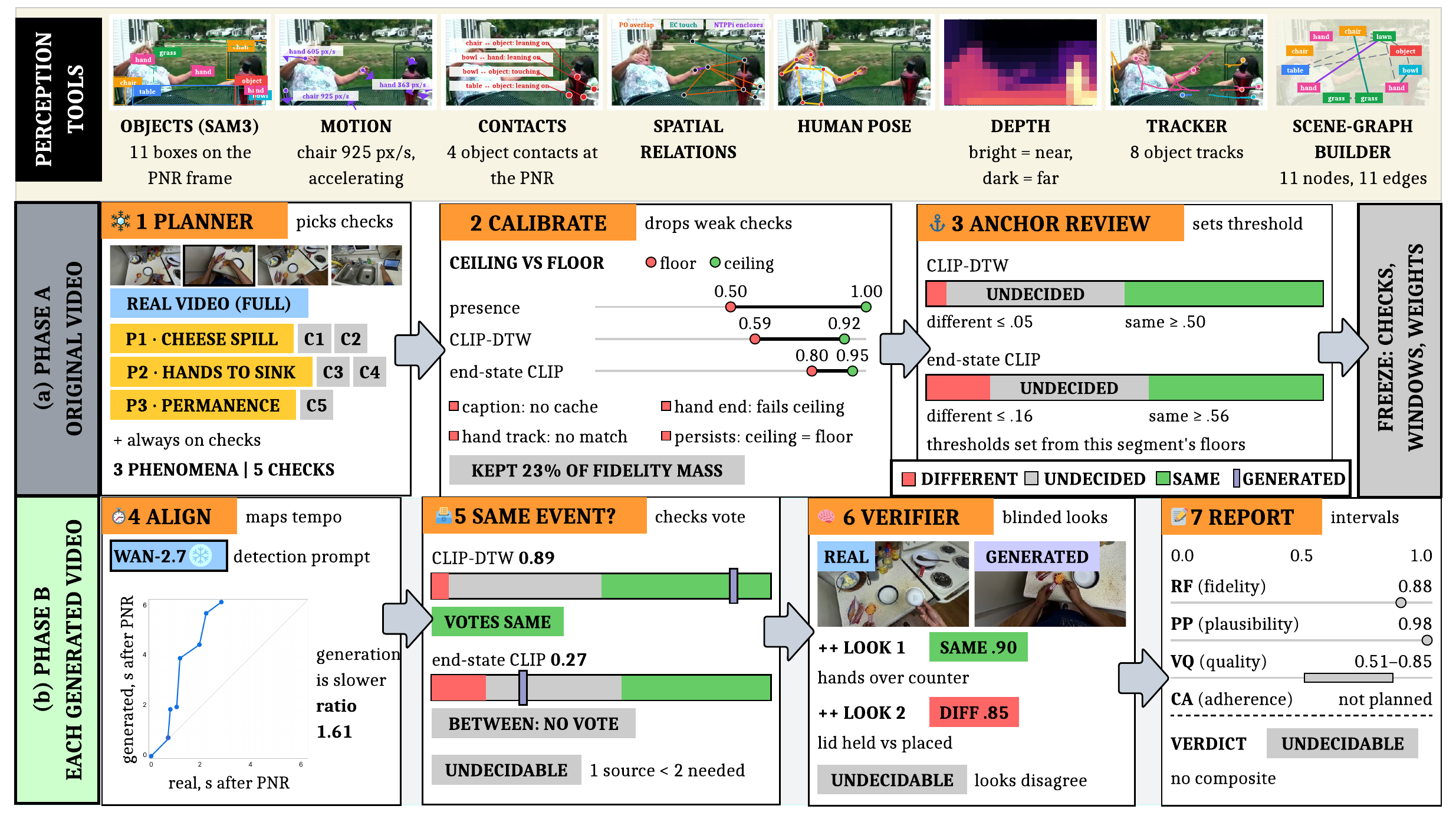}
\vspace{-8mm}
\caption{\textbf{Two-phase architecture of the agentic video evaluator.} \textbf{(a) Phase A (top):} Observing strictly the real video, the \textbf{Planner} proposes focal checks (\texttt{C1}--\texttt{C5}), calibrates them on intact vs.\ corrupted clips, sets anchor thresholds, and locks checks and weights. \textbf{(b) Phase B (bottom):} On each generated clip, the evaluator aligns post-PNR time via DTW, gathers check votes, runs two blinded \textbf{Verifier} looks, and outputs $[0, 1]$ intervals for \emph{RF}, \emph{PP}, \emph{VQ}, and \emph{CA} alongside the event verdict.}
\label{fig:evaluation}
\vspace{-6mm}
\end{figure*}

\noindent\underline{\textbf{Prompt compilation across six conditioning arms.}}
Paraphrasing $A(\mathcal{C})$ with a secondary LLM would confound semantic granularity with stylistic variance. Thus, our deterministic compiler $\Phi_\ell$ maps fixed subsets of $A(\mathcal{C})$ directly into six prompt arms (Figure~\ref{fig:anticipation}; templates in App.~\ref{app:conditioning}): (1) \textbf{Baseline} (\rev{forecast-free} control); (2) \textbf{Detection-text}, appending the early unintentional-action alert $e$ across the entire horizon; (3) \textbf{Event-text}, supplying \rev{$e$ and} the predicted \textbf{L1} global event summary across the entire horizon; (4) \textbf{Anticipation-text}, condensing \rev{$e$,} \textbf{L1}, \textbf{L2}, and \textbf{L3} into one whole-horizon prompt; (5) \textbf{Interaction-chunked}, pairing global scene context with time-windowed \textbf{L2} tuples aligned to each post-PNR chunk window $W_m$ via $c_{\mathrm{txt}}^{(m)} = \Phi_{\ell,m}(A(\mathcal{C}))$; and (6) \textbf{SG-chunked} (scene graph), extending per-chunk \textbf{L2} prompts with \textbf{L3} unlocalized scene-graph transitions and predicted end states.

\par\vspace{-2mm}

\vspace{-3mm}
\section{Technical Approach: Evaluation}
\label{sec:gen-eval}
\label{sec:agentic-eval}
\vspace{-2mm}
\label{sec:eval-method}
\noindent \underline{\textbf{Agentic Video Evaluator.}} As illustrated in Fig.~\ref{fig:evaluation}, our agentic evaluator coordinates a \textbf{Planner}, \textbf{Controller}, and \textbf{Verifier} to lock reference-video checks before inspecting any generated clip (App.~\ref{app:mllms}), where every check over vision-foundation-model perception \emph{tools} and deterministic physics \emph{skills} emits a four-way outcome (\emph{consistent}, \emph{violated}, \emph{indeterminate}, or \emph{execution-failed}) so perception failures are never mistaken for physical violations. In \textbf{Phase~A}, observing strictly the reference clip and its tool outputs, the \textbf{Planner} proposes segment-specific focal entities, events, target windows, and check weights locked prior to rollout evaluation. In \textbf{Phase~B}, the \textbf{Controller} aligns tempo via bounded Dynamic Time Warping (DTW) to evaluate contact onsets, relational transitions, and end states, while a blinded \textbf{Verifier} audits ambiguous evidence without numeric scores or generator identity to emit a three-way event verdict and $[0, 1]$ intervals across \textbf{Reference Fidelity (RF)}, \textbf{Physical Plausibility (PP)}, \textbf{Visual/Temporal Quality (VQ)}, and \textbf{Conditioning Adherence (CA)}.

\label{sec:eval-alignment}
\label{sec:eval-run}
\noindent \underline{\textbf{Human-Aligned MLLM Review Judges.}} Under our \evalprotocol{}, each candidate MLLM judge inspects a generated clip alongside $\mathcal{C}$ and $Y^*$ across ten post-PNR bins without generator or prompt labels, querying up to 13 vision tools (24 calls) and 10 four-frame strips in \emph{Interactive agent} mode to emit a \evaldataset{}-schema record (Sec.~\ref{sec:evaldataset}). Against constant baselines (including an \emph{always-unrealistic} floor), we select the open-weight backbone maximizing pre-registered human alignment $S$:
\begin{equation}
S=\tfrac15\bigl(\kappa_{\mathrm{gate}}+\kappa_{\mathrm{verdict}}+\bar\kappa_{\mathrm{axes}}+F_1^{\mathrm{anom}}+F_1^{\mathrm{div}}\bigr)
\label{eq:selection}
\end{equation}
averaging Cohen's $\kappa_{\mathrm{C}}$ on the realism gate, event verdict, and seven decided fidelity axes with span-matched $F_1^{\mathrm{anom}}$ and $F_1^{\mathrm{div}}$.

\label{sec:eval-scoring}
\label{sec:aggregate-scores}
\noindent \underline{\textbf{Evaluation Metrics and Tool-Based Scoring.}} Deploying the selected judge via Eq.~\eqref{eq:selection} (Tables~\ref{tab:human-group-a} and~\ref{tab:judge-group-b-main}), we report four pass rates (\%, $\uparrow$: \textbf{Same event} \texttt{SE}, \textbf{Semantic fidelity} over \texttt{SA}/\texttt{NH}/\texttt{SO}/\texttt{SQ}/\texttt{CL}, \textbf{Appearance fidelity} over \texttt{OI}/\texttt{SC}, and \textbf{Physical realism} \texttt{Real.}) and three per-clip defect means ($\downarrow$: \textbf{Physical anomalies} \texttt{Anom.}, maximum \textbf{Anomaly severity} $0$--$5$, and \textbf{Outcome divergences} \texttt{Div.}), ranking arms by \textbf{Same event} (ties by \textbf{Semantic fidelity}) with 95\% cluster-bootstrap intervals. Complementarily, our MLLM-free 12-axis perception scorer audits \steeraxislow{} and the scorer floor in Sec.~\ref{sec:results-plugin} (App.~\ref{app:mllms}--\ref{app:plugins}), outputting \textbf{Fidelity} (nine paired axes), \textbf{Plausibility} (three reference-free axes), and \textbf{Divergence Horizon} (s).

\vspace{-2mm}
\section{Experiments}
\label{sec:experiments}
\vspace{-2mm}
\begin{figure*}[t]
\centering
\includegraphics[width=\linewidth]{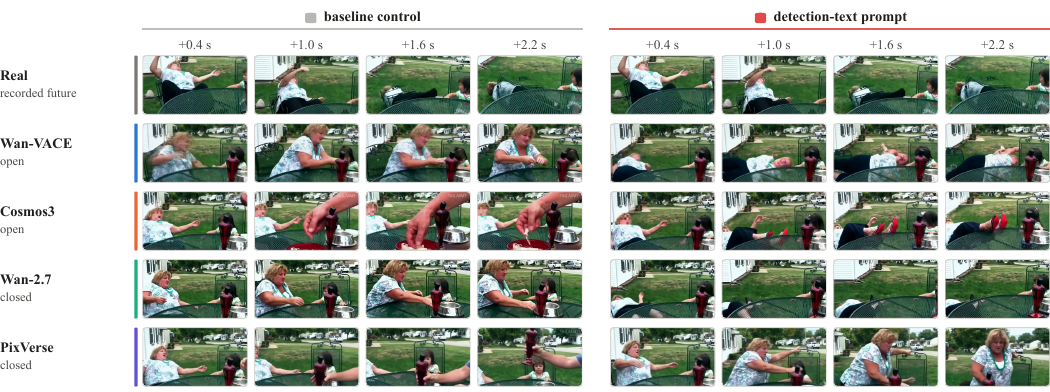}
\vspace{-5mm}
\caption{\textbf{Prediction-free vs.\ unintentional-action-conditioned rollouts (OOPS segment 12450)} at $+0.4$, $+1.0$, $+1.6$, and $+2.2$\,s under \emph{baseline/zero-shot} (left) and pre-PNR \emph{detection-text} (right) prompts (six-arm Cosmos3 rollouts in App.~\ref{app:conditioning}).}
\label{fig:cond-backbone}
\vspace{-4mm}
\end{figure*}

\label{sec:models}
\label{sec:vidgen_regimes}
\label{sec:vidgen_closed}
\label{sec:setup}
\label{sec:evidence}
\label{sec:protocol}
\noindent \underline{\textbf{Experimental Setup.}} We evaluate two open-weight V2V backbones, \textbf{Wan-VACE} (\texttt{Wan2.1-VACE-14B})\footnote{Open-weight models such as LTX-Video and CogVideoX produced degraded video-extension results in preliminary tests.} and \textbf{Cosmos3-Super}, alongside two closed video extension APIs, \textbf{Wan-2.7} and \textbf{PixVerse}, with distinct context budgets on $\mathcal{C}$ (App.~\ref{app:plugins}): at $16$\,fps, Wan-VACE conditions on up to $41$ pixel frames (${\sim}2.56$\,s) plus $x_1$ to generate $40$ new frames/chunk, whereas at $24$\,fps, Cosmos3-Super generates up to $200$ frames (${\sim}8.3$\,s) per call yet its $4\times$ causal VAE anchors only on the final $5$ pixel frames (${\approx}0.21$\,s) preceding $t_{\mathrm{pnr}}$. We evaluate three open-weight MLLMs (\texttt{Qwen3-VL-8B}, \texttt{Qwen3.8-27B}, and \texttt{Muse-Glimmer-30B}), both as released and across five Supervised Fine-Tuning (\emph{SFT-LoRA}, \emph{SFT-full}) and Group Relative Policy Optimization (\emph{GRPO-LoRA}, \emph{GRPO-full}, \emph{GRPO-full+dec}) variants trained on Physion-Eval \rev{\citep{zhang2026physioneval}} ($3 \times 6 = 18$ candidates; App.~\ref{app:mllms}).

\subsection{Results and Analysis}
\label{sec:results}

\begin{table}[t]
\centering
\scriptsize
\setlength{\tabcolsep}{3.0pt}\setlength{\aboverulesep}{0pt}\setlength{\belowrulesep}{0pt}
\renewcommand{\arraystretch}{1.22}
\caption{\textbf{MLLM judge alignment with human reviews.}}
\vspace{-3mm}
\label{tab:judge-human}
\resizebox{\linewidth}{!}{%
\begin{tabular}{@{\hspace{3pt}}l rrrr rr rrr@{\hspace{3pt}}}
\toprule
\rowcolor{tableheader}\rule{0pt}{2.4ex} & \multicolumn{4}{c}{\textbf{Original versus generated continuation}} & \multicolumn{2}{c}{\textbf{Generated clip alone}} & \multicolumn{3}{c@{\hspace{3pt}}}{\textbf{Selection score $S$ by evidence mode}} \\
\cmidrule(lr){2-5}\cmidrule(lr){6-7}\cmidrule(lr){8-10}
\rowcolor{tableheader}\textbf{Base model} & \begin{tabular}[b]{@{}r@{}}\rule{0pt}{2.4ex}\textbf{Verdict}\\\textbf{accuracy (\%)\,$\uparrow$}\end{tabular} & \begin{tabular}[b]{@{}r@{}}\rule{0pt}{2.4ex}\textbf{Verdict}\\\textbf{$\kappa_{\mathrm C}\,\uparrow$}\end{tabular} & \begin{tabular}[b]{@{}r@{}}\rule{0pt}{2.4ex}\textbf{Fidelity}\\\textbf{axes $\bar\kappa_{\mathrm C}\,\uparrow$}\end{tabular} & \begin{tabular}[b]{@{}r@{}}\rule{0pt}{2.4ex}\textbf{Divergence}\\\textbf{$F_1\,\uparrow$}\end{tabular} & \begin{tabular}[b]{@{}r@{}}\rule{0pt}{2.4ex}\textbf{Realism}\\\textbf{gate $\kappa_{\mathrm C}\,\uparrow$}\end{tabular} & \begin{tabular}[b]{@{}r@{}}\rule{0pt}{2.4ex}\textbf{Anomaly}\\\textbf{$F_1\,\uparrow$}\end{tabular} & \begin{tabular}[b]{@{}r@{}}\rule{0pt}{2.4ex}\textbf{Frames}\\\textbf{only\,$\uparrow$}\end{tabular} & \begin{tabular}[b]{@{}r@{}}\rule{0pt}{2.4ex}\textbf{One-shot}\\\textbf{summary\,$\uparrow$}\end{tabular} & \begin{tabular}[b]{@{}r@{}}\rule{0pt}{2.4ex}\textbf{Interactive}\\\textbf{agent\,$\uparrow$}\end{tabular} \\
\midrule
Qwen3-VL-8B & 77.2 & \secondmark{0.02} & -0.02 & \secondmark{0.47} & 0.00 & \secondmark{0.40} & 0.09 & 0.16 & \secondmark{0.17} \\
Qwen3.8-27B & 76.7 & \bestmark{0.09} & \bestmark{0.14} & \bestmark{0.48} & \bestmark{0.24} & \bestmark{0.44} & \bestmark{0.23} & \bestmark{0.18} & \bestmark{0.28} \\
Muse-Glimmer-30B & 78.3 & 0.00 & \secondmark{0.06} & 0.42 & \secondmark{0.07} & 0.33 & \secondmark{0.16} & \secondmark{0.17} & \secondmark{0.17} \\
\midrule
Consensus (18 variants) & 78.3 & 0.00 & 0.05 & -- & 0.00 & -- & -- & -- & -- \\
Floor (always unrealistic) & 78.3 & 0.00 & 0.00 & 0.00 & 0.00 & 0.08 & -- & -- & 0.02 \\
\bottomrule
\end{tabular}}
\end{table}

\begin{table}[t]
\centering
\scriptsize
\setlength{\tabcolsep}{2.8pt}\setlength{\aboverulesep}{0pt}\setlength{\belowrulesep}{0pt}
\renewcommand{\arraystretch}{1.20}
\caption{\textbf{\vewomooda{} human and MLLM judge results.}}
\vspace{-3mm}
\label{tab:human-group-a}
\label{tab:judge-group-a}
\resizebox{\linewidth}{!}{%
\begin{tabular}{@{\hspace{3pt}}ll rr rr rr rr rr rr rr@{\hspace{3pt}}}
\toprule
\rowcolor{tableheader} &  & \multicolumn{2}{c}{\textbf{Rank}\,$\downarrow$} & \multicolumn{2}{c}{\begin{tabular}[b]{@{}c@{}}\rule{0pt}{2.4ex}\textbf{Same}\\\textbf{event (\%)\,$\uparrow$}\end{tabular}} & \multicolumn{2}{c}{\begin{tabular}[b]{@{}c@{}}\rule{0pt}{2.4ex}\textbf{Semantic}\\\textbf{fidelity (\%)\,$\uparrow$}\end{tabular}} & \multicolumn{2}{c}{\begin{tabular}[b]{@{}c@{}}\rule{0pt}{2.4ex}\textbf{Appearance}\\\textbf{fidelity (\%)\,$\uparrow$}\end{tabular}} & \multicolumn{2}{c}{\begin{tabular}[b]{@{}c@{}}\rule{0pt}{2.4ex}\textbf{Physical}\\\textbf{realism (\%)\,$\uparrow$}\end{tabular}} & \multicolumn{2}{c}{\begin{tabular}[b]{@{}c@{}}\rule{0pt}{2.4ex}\textbf{Physical}\\\textbf{anomalies\,$\downarrow$}\end{tabular}} & \multicolumn{2}{c@{\hspace{3pt}}}{\begin{tabular}[b]{@{}c@{}}\rule{0pt}{2.4ex}\textbf{Outcome}\\\textbf{divergences\,$\downarrow$}\end{tabular}} \\
\cmidrule(lr){3-4}\cmidrule(lr){5-6}\cmidrule(lr){7-8}\cmidrule(lr){9-10}\cmidrule(lr){11-12}\cmidrule(lr){13-14}\cmidrule(lr){15-16}
\rowcolor{tableheader}\multirow{-2}{*}{\textbf{Generator}} & \multirow{-2}{*}{\textbf{Prompt}} & \textbf{Human} & \textbf{Judge} & \textbf{Human} & \textbf{Judge} & \textbf{Human} & \textbf{Judge} & \textbf{Human} & \textbf{Judge} & \textbf{Human} & \textbf{Judge} & \textbf{Human} & \textbf{Judge} & \textbf{Human} & \textbf{Judge} \\
\midrule
\multirow{2}{*}{Wan-2.7} & zero-shot & 4 & \secondmark{2} & 14.0 & \secondmark{12.3} & 36.2 & 17.2 & \secondmark{67.9} & 87.3 & \secondmark{45.3} & 32.1 & \bestmark{1.13} & 0.89 & 2.57 & \secondmark{1.49} \\
 & detection & \bestmark{1} & \bestmark{1} & \bestmark{62.2} & \bestmark{13.2} & \bestmark{64.0} & \bestmark{27.4} & \bestmark{78.3} & 93.4 & \bestmark{57.4} & 29.2 & \bestmark{1.13} & 0.85 & 2.02 & \bestmark{1.41} \\
\cmidrule{1-16}
\multirow{2}{*}{Wan-VACE} & zero-shot & 8 & 5 & 6.8 & 3.8 & 25.7 & 14.4 & 62.0 & 94.8 & 21.2 & \secondmark{47.2} & 1.38 & \secondmark{0.62} & \secondmark{1.54} & 1.67 \\
 & detection & \secondmark{2} & 4 & \secondmark{19.4} & 5.7 & 23.7 & 15.4 & 48.7 & \secondmark{95.3} & 7.7 & 36.8 & 1.67 & 0.80 & \bestmark{1.44} & 1.57 \\
\cmidrule{1-16}
\multirow{2}{*}{\rev{Cosmos3-Super}} & zero-shot & 5 & 7 & 10.0 & 1.9 & 30.4 & 12.9 & 56.2 & 78.8 & 33.3 & 38.7 & 1.60 & 0.73 & 2.00 & 1.74 \\
 & detection & 3 & 3 & 16.2 & 5.7 & \secondmark{38.2} & \secondmark{19.0} & 65.9 & \bestmark{96.7} & 43.2 & \bestmark{54.7} & \secondmark{1.29} & \bestmark{0.50} & 1.82 & 1.55 \\
\cmidrule{1-16}
\multirow{2}{*}{PixVerse} & zero-shot & 7 & 8 & 8.3 & 0.9 & 28.1 & 8.9 & 50.9 & 83.0 & 23.2 & 27.4 & 1.43 & 0.88 & 1.91 & 1.65 \\
 & detection & 6 & 6 & 9.3 & 2.8 & 23.3 & 7.7 & 49.0 & 87.3 & 22.9 & 17.0 & 1.73 & 1.06 & 2.35 & 1.65 \\
\bottomrule
\end{tabular}}
\end{table}

\label{sec:results-judge}
\label{sec:results-human}
\noindent \underline{\textbf{RQ1 \& RQ2: Human \& MLLM Judge Alignment.}} Evaluating all 18 candidate judges against human reviews on \vewomooda{} (Table~\ref{tab:judge-human}, App.~\ref{app:mllms}), released \texttt{Qwen3.8-27B} (\emph{Interactive agent}) achieves the strongest alignment ($S = 0.28$ vs.\ $0.17$) and leads every component ($\kappa_{\mathrm{gate}} = 0.24$, $F_1^{\mathrm{anom}} = 0.44$, $F_1^{\mathrm{div}} = 0.48$, $\bar\kappa_{\mathrm{axes}} = 0.14$, $\kappa_{\mathrm{verdict}} = 0.09$; \emph{Verdict accuracy} is unhighlighted below the $78.3\%$ floor). Conversely, Physion-Eval fine-tuning degrades paired-review alignment across all 18 arms (\texttt{GRPO-LoRA} drops $S$ from $0.28 \rightarrow 0.13$ on \texttt{Qwen3.8-27B} and $0.17 \rightarrow 0.07$ on the other backbones; \emph{SFT}/\emph{GRPO} $\le 0.14$), while holistic verdicts remain difficult ($141/180$ human \emph{different outcome}, $\kappa_{\mathrm{verdict}} \le 0.09$); thus, we employ Base \texttt{Qwen3.8-27B} for relative ranking.

\begin{table}[t]
\centering
\scriptsize
\setlength{\tabcolsep}{3.2pt}\setlength{\aboverulesep}{0pt}\setlength{\belowrulesep}{0pt}
\renewcommand{\arraystretch}{1.20}
\caption{\textbf{Prompt Conditioning on the \benchname{}.}}
\vspace{-3mm}
\label{tab:judge-group-b-main}
\resizebox{\linewidth}{!}{%
\begin{tabular}{@{\hspace{3pt}}rl rrrr rr@{\hspace{3pt}}}
\toprule
\rowcolor{tableheader}\textbf{Rank}\,$\downarrow$ & \textbf{Conditioning arm} & \begin{tabular}[b]{@{}r@{}}\rule{0pt}{2.4ex}\textbf{Same}\\\textbf{event (\%)\,$\uparrow$}\end{tabular} & \begin{tabular}[b]{@{}r@{}}\rule{0pt}{2.4ex}\textbf{Semantic}\\\textbf{fidelity (\%)\,$\uparrow$}\end{tabular} & \begin{tabular}[b]{@{}r@{}}\rule{0pt}{2.4ex}\textbf{Appearance}\\\textbf{fidelity (\%)\,$\uparrow$}\end{tabular} & \begin{tabular}[b]{@{}r@{}}\rule{0pt}{2.4ex}\textbf{Physical}\\\textbf{realism (\%)\,$\uparrow$}\end{tabular} & \begin{tabular}[b]{@{}r@{}}\rule{0pt}{2.4ex}\textbf{$\Delta$ Same}\\\textbf{event\,$\uparrow$}\end{tabular} & \begin{tabular}[b]{@{}r@{}}\rule{0pt}{2.4ex}\textbf{$\Delta$ Semantic}\\\textbf{fidelity\,$\uparrow$}\end{tabular} \\
\midrule
\rowcolor{subsubheader}\multicolumn{8}{@{}l}{\hspace{3pt}\textbf{\emph{Wan-VACE (full steering module)}}} \\
\midrule
\bestmark{1} & Interaction-chunked & \bestmark{7.5} & \secondmark{16.6} & \bestmark{92.5} & 33.1 & \bestmark{2.8} & \secondmark{1.9} \\
\secondmark{2} & Detection-text & \secondmark{7.1} & \bestmark{16.9} & \secondmark{92.4} & \secondmark{33.7} & \secondmark{2.4} & \bestmark{2.1} \\
3 & Event-text & \secondmark{7.1} & 15.9 & 91.4 & 33.5 & \secondmark{2.4} & 1.1 \\
4 & SG-chunked & 6.9 & 16.4 & 91.4 & 32.3 & 2.2 & 1.7 \\
5 & Anticipation-text & 6.5 & 14.9 & 90.3 & 27.8 & 1.8 & 0.2 \\
6 & Baseline$^{\dagger}$ & 4.7 & 14.7 & 91.3 & \bestmark{41.4} & -- & -- \\
\midrule
\rowcolor{subsubheader}\multicolumn{8}{@{}l}{\hspace{3pt}\textbf{\emph{\rev{Cosmos3-Super} (full steering module)}}} \\
\midrule
\bestmark{1} & Interaction-chunked & \bestmark{7.7} & \bestmark{20.0} & \bestmark{93.5} & \secondmark{44.8} & \bestmark{4.7} & \bestmark{7.2} \\
\secondmark{2} & SG-chunked & \bestmark{7.7} & \secondmark{17.8} & 89.0 & 41.8 & \bestmark{4.7} & \secondmark{5.0} \\
3 & Detection-text & \secondmark{6.3} & 17.2 & \secondmark{92.0} & \bestmark{46.2} & \secondmark{3.2} & 4.4 \\
4 & Anticipation-text & 5.9 & 16.1 & 90.0 & 37.1 & 2.8 & 3.3 \\
5 & Event-text & 5.3 & 16.6 & 91.0 & 42.4 & 2.2 & 3.8 \\
6 & Baseline$^{\dagger}$ & 3.0 & 12.8 & 78.3 & 38.7 & -- & -- \\
\bottomrule
\end{tabular}}
\end{table}

\noindent \underline{\textbf{Zero-shot vs.\ unintentional-action prompting.}}
In Table~\ref{tab:human-group-a} (\evaldataset{}, App.~\ref{app:mllms}), \emph{zero-shot} rollouts rarely capture the impending unintentional action, matching $Y^*$ on only $6.8\%$ (Wan-VACE), $8.3\%$ (PixVerse), $10.0\%$ (\rev{Cosmos3-Super}), and $14.0\%$ (Wan-2.7) of human-decided clips. Adding an unintentional-action \emph{detection} prompt lifts human same-event rates to $62.2\%$ on Wan-2.7 ($[46.3, 77.8]\%$\rev{; $+47.1$\,pp prefix-paired same-event and $+29.8$\,pp semantic gain, Table~\ref{tab:human-group-a-full}}), $19.4\%$ on Wan-VACE \rev{($+10.0$\,pp)}, $16.2\%$ on \rev{Cosmos3-Super} \rev{($+11.8$\,pp)}, and $9.3\%$ on PixVerse \rev{($+5.1$\,pp)}; Base \texttt{Qwen3.8-27B} matches this ordering, ranking Wan-2.7 first under \emph{detection} ($13.2\%$ \texttt{SE}, $27.4\%$ semantic) and \emph{zero-shot} ($12.3\%$, $17.2\%$) ahead of \rev{Cosmos3-Super}, Wan-VACE ($1.9\%$--$5.7\%$), and PixVerse ($0.9\%$--$2.8\%$).

\noindent\underline{\textbf{Insights:}} \textbf{(1)} Pretraining priors pull unguided \VDMs{} toward calm routine motion (human \texttt{SC} passes $70.6\%$), whereas synthesizing rapid falls or spills introduces physical artifacts. \textbf{(2)} Whole-clip re-rendering in Wan-2.7 distributes prompt intent across the transition better than strict frame appending. \textbf{(3)} Standalone glitch supervision on Physion-Eval induces an over-rejection bias during two-video review (App.~\ref{app:mllms}).


\label{sec:results-judged}
\label{sec:results-qual}
\noindent \underline{\textbf{RQ3: Cross-Backbone Prompt Conditioning Sensitivity.}} \rev{Across} \benchname{} segments in Table~\ref{tab:judge-group-b-main} (\rev{with standalone forecasting rising across $\textsc{ZeroShot} < \textsc{Cascaded} < \textsc{ReAgent}$ in App.~\ref{sec:results-antic};} App.~\ref{app:mllms}), both Wan-VACE and \rev{Cosmos3-Super} rank \emph{Interaction-chunked} first and \emph{Baseline} last yet differ sharply in prompt sensitivity. On \rev{Cosmos3}, every guided arm significantly outperforms \emph{Baseline}: \emph{Same event} ($3.0\%$) rises by $+2.2$ to $+4.7$ points, \emph{Semantic fidelity} ($12.8\%$) by $+3.3$ to $+7.2$ points, and \emph{Appearance fidelity} from $78.3\%$ to $89.0\%$--$93.5\%$. On Wan-VACE, guided prompts lift \emph{Same event} ($4.7\%$) by $+1.8$ to $+2.8$ points, keep \emph{Semantic fidelity} flat ($14.7\%$ vs.\ $14.9\%$--$16.9\%$), and lower \emph{Physical realism} ($41.4\% \rightarrow 27.8\%$--$33.7\%$). Qualitatively (Fig.~\ref{fig:cond-backbone}), unguided rollouts continue routine table-side motion whereas the alert triggers the backward fall.

\noindent\underline{\textbf{Insights:}} \textbf{(1)} \rev{Cosmos3}'s 5-frame (${\approx}0.21$\,s) context and structured caption fields (App.~\ref{app:conditioning}) make text its primary post-PNR driver, whereas Wan-VACE's 41-frame rolling window induces visual motion inertia. \textbf{(2)} \emph{SG-chunked} does not consistently beat \emph{Interaction-chunked} because over-predicted secondary object motion inflates extra actions.

\noindent \underline{\textbf{RQ4: \steeraxis{} vs.\ Visual Quality.}}
\label{sec:results-plugin}
Whereas prompts govern \emph{what} event unfolds past the PNR, our training-free steering module (\tediopp{}, \bachvidpp{}, \ttoname{}) governs \emph{how} attention features and noisy latents evolve. Auditing ablations with our 12-axis scorer via paired Wilcoxon signed-rank tests \citep{wilcoxon1945individual}, Benjamini--Hochberg $q$-values, and MD5 byte-identity checks (App.~\ref{app:plugins}), paired differences (e.g., $0.027$ on relations/events) stay within the byte-identical tool noise floor.

\vspace{-2mm}
\section{Conclusion, Limitations, and Future Work}
\label{sec:discussion}
\vspace{-2mm}
\noindent\textbf{(a) Conclusion:} We introduced the \benchname{} and the \evalframework{} to evaluate whether frozen \VDMs{} can continue real-world scenes past a PNR into unintentional actions. Separating our training-free \steeraxislow{} module from agentic semantic guidance along the \promptaxisfulllow{} reveals that unguided \VDMs{} preserve scene appearance yet default to routine activity. Conversely, we find that temporally aligned hierarchical MLLM semantic guidance serves as the primary lever for generating unintentional-action alike outcomes. \textbf{(b) Limitations:} Heavy commercial API costs limited our ability to generate a larger set of videos through closed-source video extension APIs (\textbf{Wan-2.7} and \textbf{PixVerse}), restricting exhaustive closed-model evaluation. \textbf{(c) Future Work:} Although our selected base agentic MLLM judge has shown some alignment with human reviewers, there can be substantial improvements \rev{alongside paired divergence-calibrated evaluators (App.~\ref{app:synthesis-challenges}). Beyond visual evaluation, our work opens several avenues for embodied anticipation: (1) lifting unlocalized scene-graph conditioning to persistent 3D world scene graphs that track occluded entities across camera motion \citep{peddi2026worldsgg}; (2) incorporating non-visual modalities such as standardized olfactory sensing \citep{france2025position,france2026chasing} to detect latent state changes (e.g., overheating or chemical spills) before visual failure cues emerge; and (3) coupling pre-PNR counterfactual rollouts with proactive augmented-reality task guidance \citep{rheault2024predictive} and point-cloud trajectory optimization \citep{xiang2024grasping} to facilitate safe robotic intervention.}
\par\vspace{-2mm}

\section{AI Use Statement}
\label{sec:ai-use}
\vspace{-2mm}
In this work, we used generative AI tools (Claude and Claude Code by Anthropic; ChatGPT and Codex by OpenAI) to clean and reformat the dataset, namely to write scripts that clean, convert, and reformat the dataset and annotation files. Additionally, we used generative AI tools to create and modify figures, charts, and tables, to edit author-written text for readability, and to identify and summarize related literature and format references. We have reviewed all AI-assisted work: that includes inspecting the data-processing scripts and checking their outputs against the source annotations, checking every figure and table against the underlying result files, and verifying each AI-suggested reference against the original publication, and curated bibliography manually.

{
    \small
    \bibliographystyle{ieeenat_fullname}
    \bibliography{cvpr2027}
}

\end{document}